\documentclass{bmvc2k}

\usepackage{amsfonts,amssymb}
\usepackage{booktabs}
\usepackage{multirow}

\title{SlotAVS: Object-Centric Audio-Visual Segmentation via Bi-Modal Slot Attention}

\addauthor{Arnav Sharma}{m24cse004@iitj.ac.in}{1}
\addauthor{Anand Mishra}{mishra@iitj.ac.in}{1}
\addauthor{Renu Rameshan}{renur@vehant.com}{2}

\addinstitution{
 Indian Institute of Technology Jodhpur\\
}
\addinstitution{
 Vehant Technologies Pvt Ltd\\
}
\runninghead{Sharma et al.}{SlotAVS}

\begin{document}

\maketitle

\begin{abstract}
Audio-Visual Segmentation (AVS) requires predicting pixel-wise masks of sounding objects by grounding acoustic signals in visual space. In multi-source scenes, concurrent sound sources often remain entangled within shared visual representations, leading to merged masks or unstable specialization when using standard alignment-focused transformer architectures. To address this, we present SlotAVS, an object-centric framework driven by competitive bi-modal slot attention. While robust foundation models provide rich semantic priors, SlotAVS shifts the paradigm from global cross-modal fusion to explicit competitive binding. By initializing discrete object slots from temporal audio sequences and enforcing strict slot-wise competition over spatial visual tokens, our method structurally mitigates the blending of distinct acoustic entities. Further, to stabilize instance-level learning, we introduce a Hungarian matching strategy that optimally aligns slots with ground-truth objects using spatial and boundary overlap. Coupled with a numerically stable, confidence-aware mask fusion mechanism, SlotAVS robustly aggregates multi-source predictions. Extensive experiments on the AVSBench dataset demonstrate the efficacy of our approach, achieving 88.6\% mIoU on the Single-Source (S4) split and 71.7\% mIoU on the Multi-Source (MS3) split. These results show that explicit competitive binding is critical for complex audio-visual scene understanding and effectively leveraging foundation-model priors without relying solely on their capacity. The code and pretrained models are publicly available at: \url{https://github.com/Arnav-Sharmaa/SlotAVS}.
\end{abstract}

\section{Introduction}
\label{sec:intro}

\begin{figure}[t]
	\centering
	\includegraphics[width=1\linewidth]{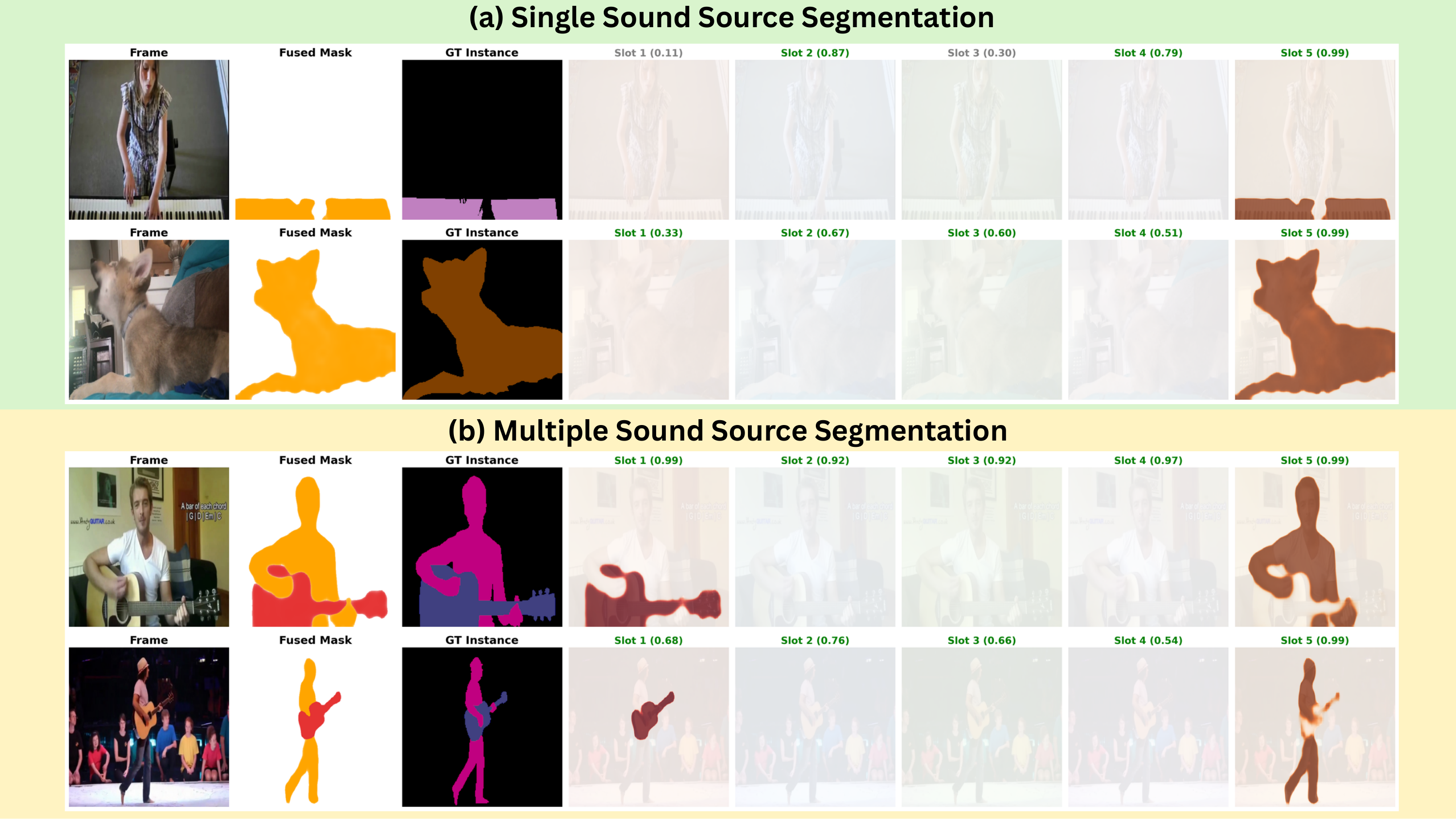}
	\caption{\textbf{Object-centric disentanglement via bi-modal slot attention.} We visualize the inner workings of SlotAVS in (a) single-source and (b) multi-source acoustic scenes. Instead of relying on a global attention map that risks entangling concurrent sounds, our architecture routes visual features into $K=5$ explicitly competitive slots. As shown, distinct sounding entities (e.g., different instruments or individuals) are strictly isolated into separate slots with high confidence scores. Unmatched slots are naturally suppressed, allowing the fused predictions to cleanly match the instance-level ground truth (GT) without hallucination or mask merging. \textbf{(Best viewed in color)}.}
	\label{fig:intro}
\end{figure}

The seamless integration of auditory and visual cues is fundamental to human perception; for example, we effortlessly associate a barking sound with a dog or a melody with a violin. Replicating this coherent audio-visual understanding is the cornerstone of robust scene perception, driving critical applications in robotics~\cite{gan2022finding,wilson2022audio}, autonomous driving~\cite{lin2024echotrack}, and anomaly detection~\cite{feng2023self, wu2025avadclip}. Among multimodal tasks, audio-visual segmentation (AVS) \cite{zhou2022audio, zhou2025audio} presents a uniquely demanding challenge. Its objective is to predict fine-grained, pixel-level segmentation masks for all sound-producing objects in a video frame, grounded strictly by the accompanying acoustic signal. Unlike audio-visual correspondence~\cite{aytar2016soundnet, arandjelovic2017look,arandjelovic2018objects} or sound source localization~\cite{afouras2020self,chen2021localizing, mo2022localizing, park2024can, tian2018audio, mo2022closer, senocak2018learning}, which typically yield coarse heatmaps, AVS demands precise cross-modal grounding. This is especially difficult in complex, multi-source scenarios where several objects produce overlapping sounds and must be spatially separated at the instance level.

To tackle this, recent AVS methods increasingly leverage powerful foundation models for rich visual and acoustic priors~\cite{luo2025tavis, liu2024bavs, liu2024annotation, mo2023av, wang2024prompting}. However, prevailing frameworks predominantly rely on global cross-modal fusion or standard multi-query transformer architectures. In these designs, standard cross-attention optimizes for alignment rather than exclusivity; because softmax normalization is applied over spatial locations, multiple queries can attend to the exact same dominant visual regions. Consequently, concurrent sound sources remain entangled within shared visual representations and create a superposition effect. Lacking structural exclusivity, the network struggles to isolate instances, resulting in merged masks or redundant predictions for spatially distinct sounding objects.

Overcoming this limitation demands a paradigm shift toward object-centric competitive binding. Rather than modeling sounding regions as a continuous, entangled activation field, objects must emerge as discrete entities that actively compete for visual ownership~\cite{locatello2020object}. To this end, we propose \textbf{SlotAVS} -- an object-centric audio-visual segmentation framework driven by \textit{Bi-Modal Slot Attention}. SlotAVS initializes discrete object slots using temporal audio token sequences extracted from a foundation acoustic model (CLAP)~\cite{wu2023large}, anchoring them to distinct acoustic events. Spatial visual tokens are then iteratively bound to these slots under strict slot-wise normalization. This introduces a zero-sum constraint: if one slot explains a visual region, the activation of other slots in that region is actively suppressed. To stabilize the assignment of these permutation-invariant slots during training, we introduce an instance-level Hungarian matching objective. This strategy optimally aligns predicted slots with ground-truth instances using robust spatial intersection-over-union (IoU) and boundary-aware Dice costs, ensuring robust instance-level specialization. Coupled with a numerically stable, probabilistic mask fusion mechanism, SlotAVS robustly aggregates these disentangled slots into coherent multi-source predictions, fully preserving the representational power of the pre-trained foundation models.

Extensive experiments across standard AVS benchmarks demonstrate consistent improvements in multi-source segmentation accuracy, robustness under complex acoustic scenes, and cross-dataset generalization, confirming that explicit competitive binding is critical for complex audio-visual scene understanding.

Our main contributions are summarized as follows: (i) We revisit audio-visual segmentation from a competitive object-centric perspective, formally identifying how spatial softmax normalization in standard queries leads to an entangled superposition of concurrent sound sources.
(ii) We introduce SlotAVS -- a novel architecture featuring bi-modal slot attention. By combining audio-guided temporal initialization with strict slot-wise spatial competition, our method explicitly separates acoustic instances and mitigates query redundancy.
(iii) We design an instance-level Hungarian matching strategy for stable supervision, optimally aligning predicted slots with ground-truth objects using rigorous mask-level metrics to enforce specialization. (iv) We seamlessly integrate frozen multimodal foundation priors within this competitive framework, isolating the algorithmic benefits of explicit instance competition and demonstrating robust generalization independent of backbone capacity.

\section{Related Work}
\label{sec:related}

\subsection{Audio-Visual Correspondence and Alignment}
\label{sec:rw-avcorr}
Learning joint representations across-modalities has historically relied on natural synchronization to drive self-supervised contrastive objectives \cite{aytar2016soundnet, arandjelovic2017look, morgado2021audio, alwassel2020self}. This paradigm scaled significantly with the advent of multimodal foundation models. Inspired by CLIP \cite{radford2021learning}, architectures such as AudioCLIP \cite{guzhov2022audioclip}, WAV2CLIP \cite{wu2022wav2clip}, CLAP \cite{elizalde2023clap,wu2023large}, and VATT \cite{akbari2021vatt} established robust, shared embedding spaces for audio and vision. More recently, unified models like ImageBind \cite{girdhar2023imagebind}, LanguageBind \cite{zhu2023languagebind}, and OmniBind \cite{wang2024omnibind} have expanded this to continuous latent spaces across heterogeneous sensors. While these foundation models provide exceptionally strong semantic priors and enable zero-shot transfer, they are fundamentally designed to optimize \textit{global} cross-modal alignment. Consequently, their global embeddings lack the spatial granularity required to disentangle multiple concurrent acoustic instances at the pixel-level.

\subsection{Audio-Visual Localization}
\label{sec:rw-avloc}
Prior to the formalization of dense segmentation, researchers focused on Audio-Visual Localization (AVL) \cite{senocak2018learning, zhao2018sound}, tracking sound sources via spatial heatmaps. Early probabilistic models \cite{kidron2005pixels, hershey1999audio} evolved into dual-stream deep networks trained via contrastive learning \cite{zhao2018sound, senocak2018learning, chen2021localizing}. Subsequent advancements refined these localization maps using hard sample mining, pseudo-labeling, and momentum encoders \cite{zhou2023exploiting, sun2023learning, senocak2023sound, mo2022closer, tian2018audio}. While recent works have extended AVL to multi-source scenarios \cite{choi2025s}, localization methods inherently yield coarse similarity heatmaps. They indicate where a sound originates but fail to delineate precise object boundaries or handle dense occlusion, necessitating the shift toward instance-level segmentation.

\subsection{Audio Visual Segmentation}
\label{sec:rw-avs}
The Audio Visual Segmentation (AVS) task, formalized by the AVSBench dataset \cite{zhou2022audio, zhou2025audio}, bridges acoustic reasoning with pixel-level visual masking. Standard approaches treat AVS as a cross-modal query response problem, projecting audio embeddings into transformer decoders to generate dense masks. For instance, AVSegformer \cite{gao2024avsegformer} and AVESformer \cite{wang2024avesformer} amplify auditory queries to guide visual spatial temporal attention. To better localize sounding objects in complex environments, subsequent works have focused on query dependency modeling (CATR \cite{li2023catr}), audio aware query enhancement (AuTR \cite{liu2023audio}), and quantization based semantic decomposition (QDFormer \cite{li2024qdformer}).

As the field progressed, researchers introduced advanced learning paradigms to refine cross-modal alignment. These include consistency queried cross-modal attention \cite{lv2025consistency}, multi order bilateral relations \cite{yang2024cooperation}, bidirectional generation \cite{hao2024improving}, as well as contrastive~\cite{gong2025complementary} and counterfactual learning~\cite{zha2025implicit} frameworks to mitigate bias and improve training stability. To address optimization bottlenecks in end to end training, recent methods have also proposed progressive two stage learning strategies alongside adaptive query generators~\cite{ma2024stepping}. Concurrently, there has been a push to rethink the architectural formulation itself, ranging from state space models like AVS-Mamba~\cite{gong2025avs} and vision centric transformers~\cite{huang2025revisiting}, to methods that explicitly explore instance associations~\cite{chen2024unraveling}, dynamic audio semantic elimination~\cite{liu2025dynamic}, and robust audio guided visual convergent alignment~\cite{liu2025robust}. Most recently, the field has rapidly integrated powerful multimodal foundation models to enhance zero shot generalization and mask fidelity. Approaches such as BAVS~\cite{liu2024bavs} bootstrap segmentation using integrated foundation knowledge, while methods such as TAViS~\cite{luo2025tavis} and AV-SAM~\cite{mo2023av} leverage text as a semantic bridge or adapt the Segment Anything Model (SAM~\cite{kirillov2023segment,ravi2024sam}) to align frozen audio visual priors.

Despite these advances, existing frameworks share the following critical structural limitation: they rely on alignment driven cross attention that normalizes over spatial dimensions. This mathematically permits multiple latent queries to collapse onto the exact same visual regions. Lacking explicit competition, concurrent sound sources remain entangled within shared visual representations and create a \textit{superposition effect}. SlotAVS overcomes this via competitive object-centric binding, structurally prohibiting spatial overlap between discrete acoustic entities.

\subsection{Object-Centric Learning and Slot Attention}
\label{sec:rw-ocl}
Object-centric learning parses scenes into discrete entities rather than continuous feature maps. Slot attention~\cite{locatello2020object} achieves this by binding spatial features to learnable slots via competitive normalization (applying a softmax over the slot dimension). Originally designed for static images, this paradigm has been extended to audio~\cite{reddy2023audioslots} and multimodal reasoning~\cite{han2025slotfusion,xu2024slot}. However, within the audio-visual localization domain, joint slot mechanisms remain restricted to coarse foreground-background localization~\cite{kim2025improving}. Consequently, existing multimodal applications lack the explicit structural mechanisms and supervised training paradigms required for dense, instance-level segmentation of complex acoustic mixtures.
SlotAVS advances this paradigm through bi-modal slot attention, which anchors slots using temporal audio sequences rather than random initialization. To resolve the instability of permutation-invariant slot matching during supervised training, we introduce an instance-level Hungarian matching objective based on rigorous spatial overlap (IoU and Dice). This effectively adapts the disentanglement capabilities of slot attention for dense, multi-source AVS.
\section{Methodology}
\label{sec:method}

\begin{figure*}[t]
	\centering
	\includegraphics[width=1\textwidth]{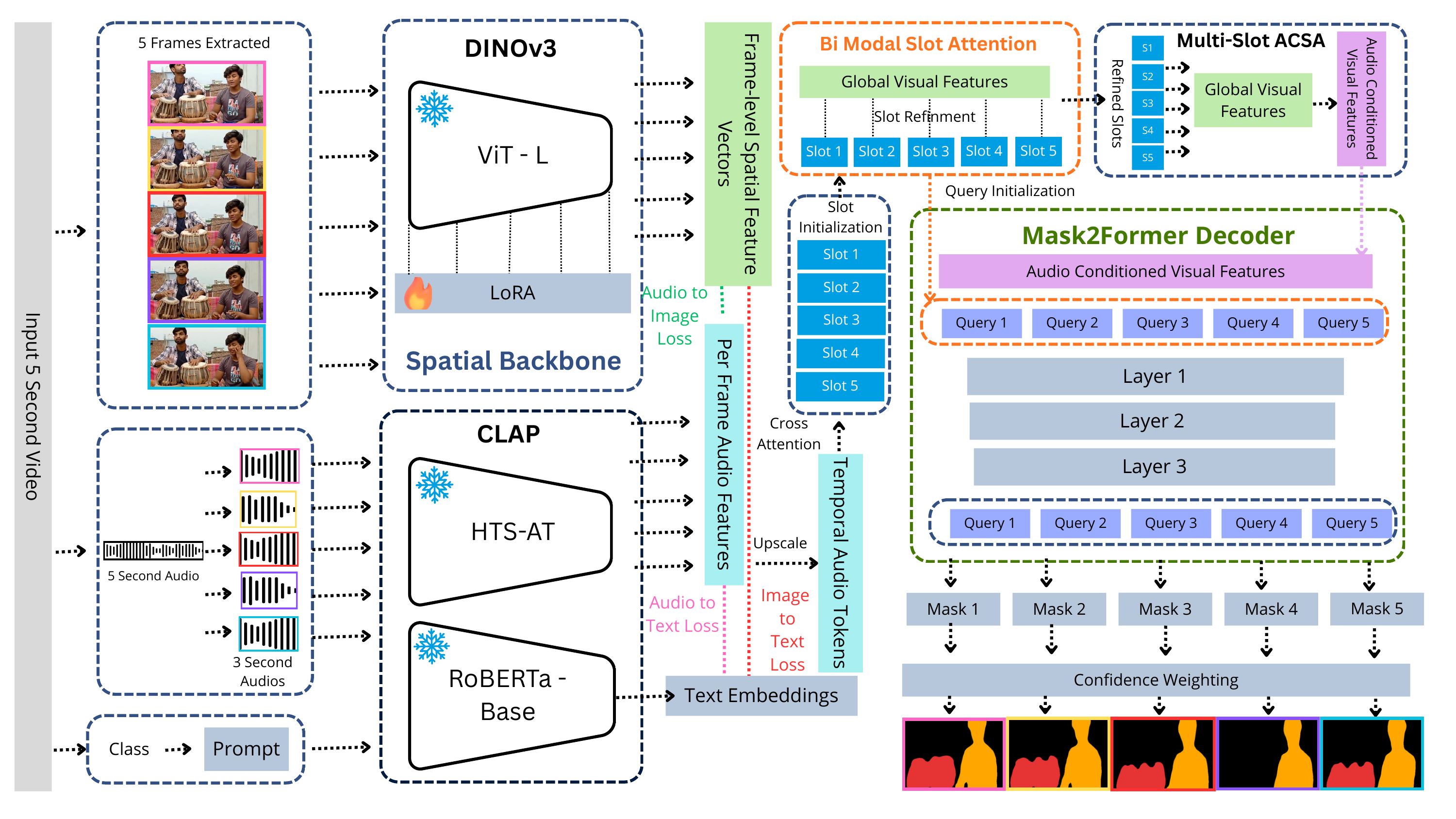}
	\caption{Overall architecture of SlotAVS. The framework leverages frozen foundation models (DINOv3 and CLAP) to extract visual and audio representations. A bi-modal slot attention module performs competitive cross-modal binding to establish object-centric slots. These slots condition the spatial features via Multi-Slot ACSA and act as queries in a mask decoder, followed by instance aware Hungarian matching, probabilistic fusion, and tri modal Contrastive Feature Distillation (CFD).}
	\label{fig:overall-arch}
\end{figure*}

\subsection{Overview}
\label{sec:overview}
The proposed SlotAVS aims to generate explicit, instance level segmentation masks for multiple concurrent sounding entities. Instead of relying on global cross-modal alignment, which often entangles spatially overlapping or semantically similar sound sources, we formulate audio visual segmentation as a problem of competitive binding.

Given a video frame and its accompanying audio waveform, we extract high resolution spatial visual tokens $\mathbf{F}_{vis} \in \mathbb{R}^{N \times D}$ using a frozen vision foundation model (we use DINOv3~\cite{simeoni2025dinov3} adapted via LoRA~\cite{hu2022lora}). Concurrently, an audio foundation model (we use CLAP~\cite{wu2023large}) extracts both a global audio embedding $\mathbf{f}_{audio} \in \mathbb{R}^{D}$ and a sequence of temporal audio tokens $\mathbf{T}_{audio} \in \mathbb{R}^{T_{aud} \times D}$. The SlotAVS framework operates in four core stages: (i) initializing object slots from temporal audio semantics and refining them via competitive cross-modal binding; (ii) modulating spatial features through multi-slot audio-conditioned spatial attention (ACSA); (iii) decoding mask probabilities using a slot based transformer decoder; and (iv) supervising the model via instance level Hungarian matching, probabilistic mask fusion, and tri-modal feature distillation.

\begin{figure}[!t]
	\centering
\includegraphics[width=1\textwidth]{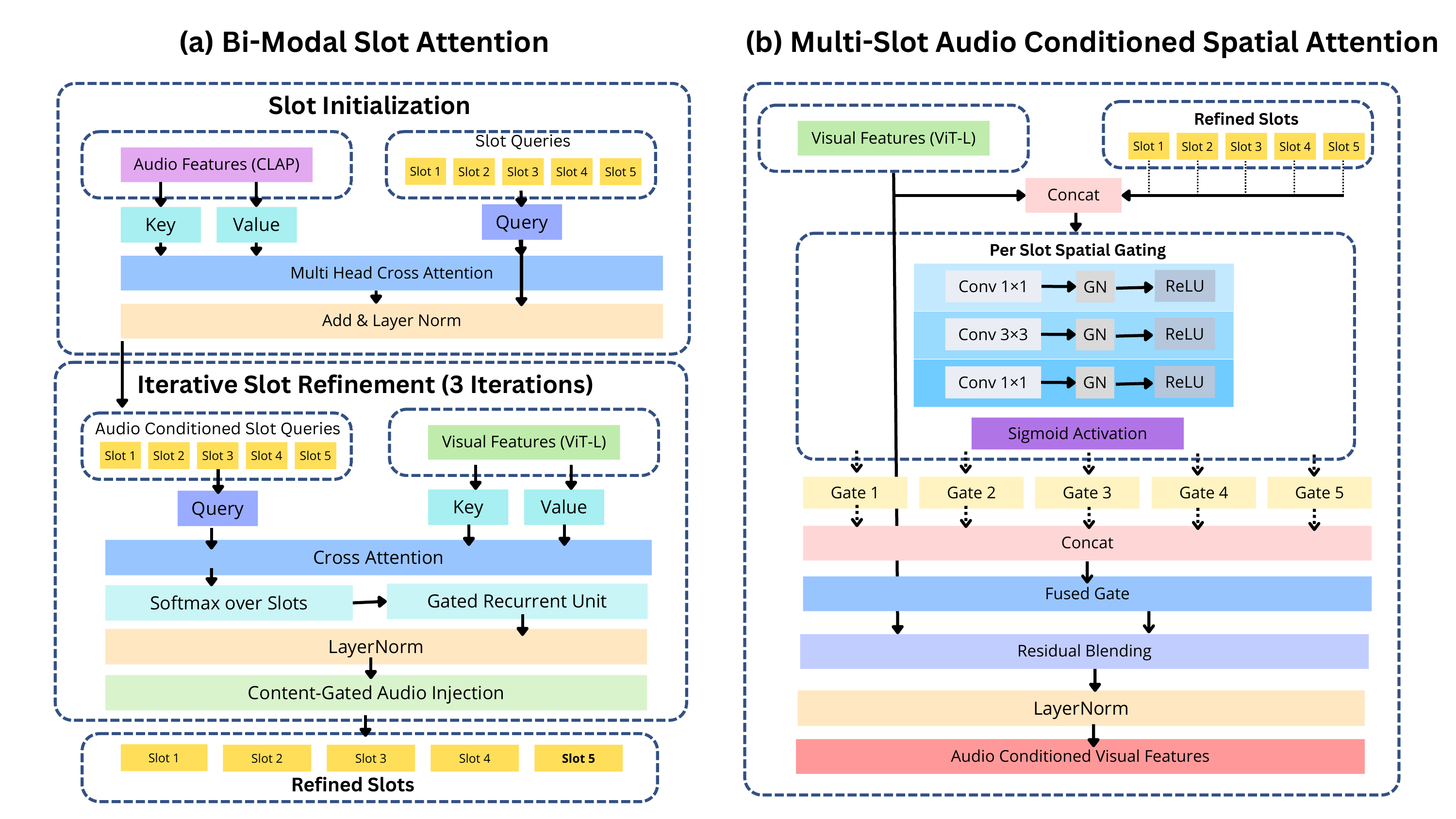}
\caption{Detailed schematic of the (a) Bi-modal slot attention and (b) Multi-slot audio-conditioned spatial attention (ACSA) modules. Note the slot wise softmax operation that encourages zero sum competitive binding among the $K$ slots for visual tokens. \textbf{(Best viewed in color)}.}
	\label{fig:internal-modules}
\end{figure}

\subsection{Bi-Modal Slot Attention}
\label{sec:bimodal-slot}
To mitigate the entanglement of concurrent sound sources, we introduce a Bi Modal Slot Attention mechanism. While existing query based segmentation architectures often rely on static learnable embeddings that are fixed after training, our method utilizes dynamic slots that emerge from iterative competitive attention conditioned on audio features.

Let $\mathbf{S}^{(0)} \in \mathbb{R}^{K \times D}$ denote a set of $K$ learnable slot queries, and $\mathbf{T}_{audio}$ represent the extracted audio token sequence. We initialize the audio aware slots by cross attending directly to this sequence:
\begin{equation}
	\mathbf{S}^{(init)} = \text{LayerNorm}\left( \mathbf{S}^{(0)} + \text{MHA}(\mathbf{S}^{(0)}, \mathbf{T}_{audio}, \mathbf{T}_{audio}) \right),
\end{equation}
where $\text{MHA}$ denotes Multi-Head Attention. This tokenized cross attention allows different slots to independently anchor to distinct temporal acoustic features (e.g., a sudden drum transient versus a continuous harmonic sustain).

Following initialization, we iteratively bind the spatial visual tokens to the slots over $T_{iter}$ iterations. At iteration $t$, we project the slots into queries $\mathbf{q} = \mathbf{S}^{(t-1)}\mathbf{W}_q$, and the visual features into keys $\mathbf{k} = \mathbf{F}_{vis}\mathbf{W}_k$ and values $\mathbf{v} = \mathbf{F}_{vis}\mathbf{W}_v$. The affinity between slots and spatial tokens is computed using scaled dot-product attention. Crucially, to introduce explicit competition, the attention weights are normalized using a softmax over the $K$ slots rather than over the spatial tokens:
\begin{equation}
	A_{k, n} = \frac{\exp(\frac{\mathbf{q}_k \cdot \mathbf{k}_n^T}{\sqrt{D}})}{\sum_{j=1}^K \exp(\frac{\mathbf{q}_j \cdot \mathbf{k}_n^T}{\sqrt{D}})}.
\end{equation}
This formulation strongly discourages spatial overlap between discrete acoustic entities. Each spatial pixel is forced to distribute its probability mass among the $K$ competing slots, mitigating the issue of multiple latent queries collapsing onto the same visual region. 

To update the slot states, the attention weights are spatially normalized and multiplied by the visual values $\mathbf{v}$ to produce an aggregated visual update $\mathbf{U} \in \mathbb{R}^{K \times D}$. These updates are processed through a gated recurrent unit (GRU~\cite{cho2014learning}), utilizing $\mathbf{S}^{(t-1)}$ as the hidden state, to yield the updated state $\mathbf{S}^{(t)}_{gru}$. To prevent the slots from drifting away from their original acoustic context during iterative refinement, we perform a content gated re-injection of the global audio embedding $\mathbf{f}_{audio}$:
\begin{equation}
	\mathbf{S}^{(t)} = \mathbf{S}^{(t)}_{gru} + \sigma\big(\text{MLP}([\mathbf{S}^{(t)}_{gru}, \mathbf{f}_{audio}])\big) \odot \mathbf{f}_{audio},
\end{equation}
where $[\cdot, \cdot]$ denotes concatenation and $\sigma$ is the sigmoid function. Because the MLP gate is conditioned on the distinct hidden state $\mathbf{S}^{(t)}_{gru}$ of each individual slot, it extracts only the slot-relevant acoustic features. This ensures the refined slots remain strictly grounded to their specific sound source while absorbing visual features, preventing them from converging toward the average global audio.

\subsection{Multi-Slot Audio-Conditioned Spatial Attention (ACSA)}
\label{sec:acsa}
To project the disentangled latent slots back into the spatial domain, we design a multi-slot Audio-conditioned spatial attention (ACSA) module. Instead of computing a single global spatial attention map, we compute $K$ distinct spatial attention gates $\mathbf{G}_k \in [0, 1]^{H \times W}$, generated via a convolutional network acting on the concatenated spatial features and the respective slot vector $\mathbf{S}_k$. 

To fuse these gates into a comprehensive spatial memory, we employ a winner takes all max-pooling operation across the $K$ slots:
\begin{equation}
	\mathbf{G}_{fused} = \max_{k \in \{1 \dots K\}} \mathbf{G}_k.
\end{equation}
The visual features are modulated by this fused gate and combined with the original spatial features via a learned residual weight $\alpha$, yielding slot conditioned features $\hat{\mathbf{F}}_{vis} = (1 - \sigma(\alpha)) \mathbf{F}_{vis} + \sigma(\alpha) (\mathbf{G}_{fused} \odot \mathbf{F}_{vis})$. While $\mathbf{G}_{fused}$ provides the decoder with a rich, audio filtered global spatial context, the object-centricity of the framework is strictly preserved by the decoder queries. Specifically, unlike standard decoders that rely on static learned queries, our Mask2Former style decoder directly utilizes the $K$ fully evolved, disentangled slots $\mathbf{S}^{(T_{iter})}$ as dynamic queries to decode $K$ independent mask probabilities $\hat{\mathbf{M}} \in [0, 1]^{K \times H \times W}$.

\subsection{Instance Aware Supervision}
\label{sec:supervision}
Because the output ordering of the $K$ slots is permutation invariant, applying standard pixel wise loss is ill-posed. AVSBench natively provides instance level annotations via color coded RGB masks. We extract $N$ binary ground truth masks $\mathbf{Y} = \{\mathbf{y}_1, \dots, \mathbf{y}_N\}$ from these unique non-black colors (where $N=1$ for single source, and $N \ge 2$ for multi-source scenes). We formulate the instance level supervision as a bipartite matching problem, finding the optimal assignment $\pi$ between the $K$ predicted slot masks and $N$ ground truth instances using the Hungarian algorithm~\cite{kuhn1955hungarian,carion2020end}.

While the matching procedure follows standard set prediction frameworks, our formulation differs fundamentally because the matched elements are dynamically evolved, acoustically grounded slots. To ensure precise spatial overlap between predictions and ground truth, the cost matrix $\mathcal{C} \in \mathbb{R}^{K \times N}$ is computed using combined region overlap metrics:
\begin{equation}
	\mathcal{C}(k, n) = \lambda_{iou} \big(1 - \text{IoU}(\hat{\mathbf{M}}_k, \mathbf{y}_n)\big) + \lambda_{dice} \big(1 - \text{Dice}(\hat{\mathbf{M}}_k, \mathbf{y}_n)\big)
\end{equation}
Once the optimal bipartite matching is established, valid slot instance pairs are supervised using Binary Cross-Entropy (BCE) and Dice losses. Crucially, the remaining unmatched slots ($k \notin \pi$) are actively penalized with a no-object loss, forcing them to predict $\mathbf{0}^{H \times W}$. 

\subsection{Training Objective}
\label{sec:objective}
To dynamically model a variable number of sound sources, we introduce a Slot Confidence Head that predicts a scalar $c_k \in [0, 1]$ per slot. The confidence supervision is twofold. First, following the Hungarian assignment, matched slots receive a binary target of 1 and unmatched slots a target of 0, supervised via individual BCE losses. Second, to act as an aggregate count regularizer, the sum of all slot confidences $\sum_{k=1}^K c_k$ is supervised via Mean Squared Error (MSE) against the true instance count $N$.

To generate the final multi-source segmentation mask required for AVS evaluation, we fuse the $K$ masks using a probabilistic OR formulation. This ensures all active slots receive proportional gradient updates. The slot probabilities are clamped to prevent numerical underflow, yielding the fused global mask:
\begin{equation}
	\mathbf{P}_{global} = 1 - \prod_{k=1}^K \big(1 - \text{clamp}(c_k \hat{\mathbf{M}}_k, \, 0.0, \, 0.95)\big).
\end{equation}
$\mathbf{P}_{global}$ is subsequently supervised against the combined ground truth mask using BCE, Dice, and boundary distance-over-union (BoundaryDoU) losses.

Furthermore, we explicitly encourage slot orthogonality to prevent multiple slots from collapsing onto the loudest acoustic source. This is achieved via two diversity regularizations on the $\ell_2$-normalized slot vectors $\tilde{\mathbf{S}}$: a squared cosine similarity penalty for pairwise orthogonality, and a Determinantal Point Process (DPP~\cite{kulesza2012determinantal}) loss to maximize the global volume spanned by the slots:
\begin{equation}
	\mathcal{L}_{div} = \frac{1}{K(K-1)} \sum_{i \neq j} (\tilde{\mathbf{S}}_i \cdot \tilde{\mathbf{S}}_j)^2 - \lambda_{DPP} \log \det(\tilde{\mathbf{S}} \tilde{\mathbf{S}}^T + \epsilon \mathbf{I}).
\end{equation}

Finally, to guarantee that the visually disentangled segments remain strictly coupled to their acoustic semantic identity, we apply a Contrastive Feature Distillation (CFD) loss. During training, we synthesize text prompts describing the ground truth audio classes (e.g., ``the sound of a guitar'') to extract a global text embedding $\mathbf{f}_{text}$ via the CLAP text encoder. Importantly, this text modality is utilized exclusively as an auxiliary training objective to shape the latent space and is entirely discarded during inference, ensuring fair comparisons with standard audio visual baselines. We compute a mask pooled visual embedding $\mathbf{v}_{pool}$ by applying the global mask $\mathbf{P}_{global}$ as soft attention weights over $\mathbf{F}_{vis}$. The CFD loss enforces a tri modal alignment by minimizing the cosine distance between the audio, text, and visual projections in a shared latent space:
\begin{equation}
\begin{split}
	\mathcal{L}_{CFD} &= \lambda_{a2t}\big(1 - \cos(\mathbf{f}_{audio}, \mathbf{f}_{text})\big) \\
	&\quad + \lambda_{v2t}\big(1 - \cos(\mathbf{v}_{pool}, \mathbf{f}_{text})\big) \\
	&\quad + \lambda_{a2v}\big(1 - \cos(\mathbf{f}_{audio}, \mathbf{v}_{pool})\big).
\end{split}
\end{equation}
The total objective is the weighted sum of the global mask losses, the Hungarian matched instance losses, the confidence BCE and MSE count losses, the slot diversity regularizations, and the CFD tri modal alignment loss.

\section{Experiments and Results}
\subsection{Datasets and Evaluation Metrics}
\label{sec:datasets}

\textbf{Datasets:} We evaluate SlotAVS primarily on the AVSBench~\cite{zhou2022audio} dataset, the standard benchmark for audio-visual segmentation. AVSBench-Object comprises two subsets: Single Sound Source Segmentation (S4), containing 4,932 videos, and Multiple Sound Source Segmentation (MS3), containing 424 videos. While the S4 split tests basic cross-modal alignment, the MS3 subset serves as the critical testbed for our core architectural claims, as it explicitly evaluates a model's ability to isolate and disentangle concurrent sounding objects. We also report performance on the AVSBench-Semantic (AVSS)~\cite{zhou2025audio} split to assess category-aware segmentation capabilities. Furthermore, to demonstrate the cross-dataset generalization of our learned object-centric representations, we evaluate our model on the recently introduced AVATAR~\cite{choi2025s} benchmark for video-centric audio-visual localization. Derived from VGGSound~\cite{chen2020vggsound}, AVATAR encompasses diverse and complex acoustic conditions, explicitly testing model robustness across four fine-grained evaluation scenarios: single-sound, mixed-sound, multi-entity, and off-screen.

\noindent\textbf{Evaluation Metrics:} Following the established AVS protocol, we quantify segmentation performance using two primary metrics: the Jaccard Index (mean Intersection over Union, denoted as $\mathcal{M}_J$) to measure overall region similarity, and the F-score ($\mathcal{M}_F$) to evaluate mask boundary and contour accuracy. For the video-centric audio-visual localization benchmarks, we measure spatial grounding accuracy using consensus Intersection over Union (cIoU) and the Area Under the Curve (AUC). Finally, to evaluate the model's robustness against hallucination in off-screen scenarios, we report the True Negative (TN) rate, which quantifies the model's ability to correctly predict empty masks when a sound source is acoustically present but absent from the visual field.

\subsection{Implementation Details}
\label{sec:implementation}
\textbf{Architecture Configuration:} We use a frozen DINOv3-Large visual backbone, injecting trainable LoRA modules ($r=16, \alpha=32$, dropout $0.05$) exclusively into the attention QKV projections. Acoustic and textual features are extracted via a frozen CLAP encoder. The bi-modal slot attention module utilizes $K=5$ learnable slots and $T_{iter}=3$ refinement iterations. Inputs are standardized to $T=5$ frames. While our primary model operates at a $512 \times 512$ spatial resolution, we also evaluate variants at $224 \times 224$ and $1024 \times 1024$ to assess the balance between computational scalability and spatial granularity.

\noindent\textbf{Training Protocol:} SlotAVS is implemented in PyTorch with Automatic Mixed Precision (AMP). We optimize the network using AdamW (base learning rate $5 \times 10^{-4}$, weight decay $0.01$) with a cosine annealing schedule, a $3$-epoch linear warmup, and gradient clipping at $1.0$. Loss coefficients are empirically set to: $\lambda_{bce}=1.0$, $\lambda_{dice}=1.5$, $\lambda_{dou}=0.2$, $\lambda_{empty}=0.5$, $\lambda_{conf}=0.5$, and $\lambda_{div}=0.1$. The tri modal CFD loss ($\lambda_{a2t}=\lambda_{v2t}=0.5$, $\lambda_{a2v}=0.3$) activates after epoch $3$. 

\noindent\textbf{Hardware and Computational Profile:} The model is trained on $2$ NVIDIA A30 GPUs using a per GPU batch size of $2$ and $16$ gradient accumulation steps (effective batch size $64$). By freezing the heavy foundation backbones, SlotAVS requires only $30.0$M trainable parameters out of a total $486.6$M. To comprehensively evaluate computational expense, we profiled the framework across input resolutions. At our primary $512 \times 512$ resolution, the model requires $3446.5$ GFLOPs and consumes $2.12$ GB of peak memory per frame, yielding an inference latency of $403.7$ ms ($2.48$ FPS). While stepping down to $224 \times 224$ significantly reduces the computational burden to $769.3$ GFLOPs, the drastically reduced visual token count ($196$ vs $1024$ tokens) provides insufficient spatial density for accurate slot competition, resulting in a drop below baseline performance. Conversely, scaling to $1024 \times 1024$ improves boundary sharpness but exponentially increases cost to $13379.3$ GFLOPs. Consequently, $512 \times 512$ is adopted as the optimal operating resolution. A comprehensive profiling table detailing all computational metrics is provided in the supplementary material.

\subsection{Results}
\label{sec:results}
\noindent\textbf{Audio-Visual Segmentation on AVSBench:} 
Table~\ref{tab:results-avsbench} presents the quantitative comparison of SlotAVS against recent baseline methods on the AVSBench dataset. We evaluate our architecture across multiple spatial resolutions ($224 \times 224$, $512 \times 512$, and $1024 \times 1024$). On the Single Sound Source Segmentation (S4) benchmark, SlotAVS demonstrates highly competitive performance. Given that S4 primarily tests fundamental cross-modal alignment where a single dominant sound source is present, global fusion approaches equipped with heavy foundation backbones naturally perform well. 

However, the critical differentiator of our framework emerges in the Multiple Sound Source Segmentation (MS3) setting. In scenarios where multiple acoustic events occur concurrently, baseline architectures relying on standard cross-attention suffer from feature entanglement, resulting in a significant performance drop compared to their S4 baselines. By shifting to explicit competitive binding via Bimodal Slot Attention, SlotAVS successfully preserves object-level disentanglement. Remarkably, SlotAVS operating at a $512 \times 512$ resolution outperforms prior top-performing models operating at double the spatial resolution ($1024 \times 1024$) on the challenging MS3 split. At our maximum $1024 \times 1024$ resolution, SlotAVS surpasses all evaluated baselines, yielding exceptional $\mathcal{M}_J$ and $\mathcal{M}_F$ scores that firmly validate the necessity of structural zero-sum constraints for multisource separation.
\begin{table}[t]
	\caption{Quantitative comparison on S4 and MS3. $\mathcal{M}_J$ and $\mathcal{M}_F$ denote region similarity and contour accuracy, respectively. Baseline methods are sorted by MS3 $\mathcal{M}_J$. AVSS reports $\mathcal{M}^{1}_{J}$.}
	\label{tab:results-avsbench}
	\centering
	\small
	\setlength{\tabcolsep}{3pt} 
	\renewcommand{\arraystretch}{1.05}
	\begin{tabular*}{\linewidth}{@{\extracolsep{\fill}}llcccccc@{}}
		\toprule
		Models & Backbone & Size & \multicolumn{2}{c}{S4} & \multicolumn{2}{c}{MS3} & AVSS \\
		\cmidrule(lr){4-5}\cmidrule(lr){6-7}
		&  &  & $\mathcal{M}_J$ & $\mathcal{M}_F$ & $\mathcal{M}_J$ & $\mathcal{M}_F$ & $\mathcal{M}^{1}_{J}$ \\
		\midrule
		AV-SAM~\cite{mo2023av}            & ViT-H  & 1024 & 40.5 & 0.566 & --   & --    & - \\
		SAM~\cite{kirillov2023segment}    & ViT-H  & 1024 & 55.1 & 0.739 & 54.0 & 0.638 & - \\
		AVSBench~\cite{zhou2025audio}     & PVT-v2 & 224  & 78.7 & 0.879 & 54.0 & 0.645 & 29.7 \\
		ECMVAE~\cite{mao2023multimodal}   & PVT-v2 & 224  & 81.7 & 0.901 & 57.8 & 0.708 & - \\
		DiffusionAVS~\cite{mao2025contrastive} & PVT-v2 & 224  & 81.4 & 0.902 & 58.2 & 0.709 & - \\
		AVSegFormer~\cite{gao2024avsegformer}  & PVT-v2 & 224  & 82.1 & 0.899 & 58.4 & 0.693 & 36.7 \\
		BAVS~\cite{liu2024bavs}           & PVT-v2 & 224  & 82.0 & 0.886 & 58.6 & 0.655 & 32.6 \\
		CATR~\cite{li2023catr}            & PVT-v2 & 224  & 81.4 & 0.896 & 59.0 & 0.700 & - \\
		COMBO~\cite{yang2024cooperation}  & PVT-v2 & 224  & 84.7 & 0.919 & 59.2 & 0.712 & 42.1 \\
		UFE-AVS~\cite{liu2024audio}       & PVT-v2 & 224  & 83.2 & 0.904 & 59.5 & 0.676 & - \\
		QDFormer~\cite{li2024qdformer}    & Swin-T & 224  & 79.5 & 0.882 & 61.9 & 0.661 & - \\
		GAVS~\cite{wang2024prompting}     & ViT-B  & 1024 & 80.1 & 0.902 & 63.7 & 0.774 & - \\
		SAMA-AVS~\cite{liu2024annotation} & ViT-H  & 1024 & 83.2 & 0.901 & 66.9 & 0.754 & - \\
        AVSStone~\cite{ma2024stepping}    & Swin-B & 224  & 83.2 & 0.913 & 67.3 & 0.776 & 48.5 \\
        CAVP~\cite{chen2024unraveling}    & PVT-v2 & 224  & \underline{87.3} & \textbf{0.936} & 67.3 & 0.781 & 48.6 \\
        VCT~\cite{huang2025revisiting}    & Swin-B & 384  & 86.2 & 0.934 & 67.6 & 0.814 & \underline{51.2} \\
		TAViS~\cite{luo2025tavis}         & ViT-L  & 224  & 84.8 & 0.912 & 68.2 & 0.759 & 44.2 \\
		TAViS~\cite{luo2025tavis}         & ViT-L  & 1024 & 87.0 & 0.926 & \underline{71.2} & \underline{0.796} & - \\
		\midrule
		SlotAVS (Ours)                    & ViT-L  & 224  & 84.5 & 0.904 & 64.4 & 0.765 & 45.6 \\
        SlotAVS (Ours)                    & ViT-L  & 512  & 87.1 & 0.923 & 71.4 & 0.811 & 65.47 \\
		SlotAVS (Ours)                    & ViT-L  & 1024 & \textbf{88.6} & \underline{0.935} & \textbf{71.7} & \textbf{0.815} & \textbf{69.99} \\
		\bottomrule
	\end{tabular*}
\end{table}

\noindent\textbf{Cross-Dataset Generalization on AVATAR:}
To ensure that our competitive slots learn generalized acoustic-visual semantics rather than overfitting to AVSBench biases, we evaluate zero-shot transfer on the AVATAR video-centric localization benchmark. As shown in Table~\ref{tab:results-avloc}, SlotAVS demonstrates robust performance across all four demanding subsets. Specifically, in the Mixed-sound and Multi-entity splits, our object-centric disentanglement enables precise localization where traditional single-query methods typically fail. Furthermore, our probabilistic mask fusion and explicit slot confidence modeling ensure high True Negative (TN) rates on the Off-screen subset, successfully preventing the hallucination of objects when sounds originate outside the visual field.

\begin{table}[t]
	\caption{\textbf{Quantitative evaluation on the AVATAR audio-visual localization benchmark.} We compare SlotAVS against localization baselines across four distinct acoustic scenarios: Single-sound, Mixed-sound, Multi-entity, and Off-screen. Performance is measured using Consensus Intersection over Union (CIoU), Area Under Curve (AUC), and True Negative (TN) rates. Higher values indicate better performance.}
	\label{tab:results-avloc}
	\centering
	\scriptsize
	\setlength{\tabcolsep}{1pt}
	\renewcommand{\arraystretch}{1.0}
	\begin{tabular*}{\linewidth}{@{\extracolsep{\fill}}lccccccc@{}}
		\toprule
		& \multicolumn{2}{c}{(1) Single-sound} & \multicolumn{2}{c}{(2) Mixed-sound} & \multicolumn{2}{c}{(3) Multi-entity} & (4) Off-screen \\
		\cmidrule(lr){2-3}\cmidrule(lr){4-5}\cmidrule(lr){6-7}\cmidrule(lr){8-8}
		Method & CIoU(\%) & AUC(\%) & CIoU(\%) & AUC(\%) & CIoU(\%) & AUC(\%) & TN(\%) \\
		\midrule
		SLAVC(144k)~\cite{mo2022closer}     & 9.07  & 10.60 & 6.31  & 7.88  & 6.41  & 7.96  & 96.46 \\
		EZ-VSL(10k)~\cite{mo2022localizing}     & 9.66  & 11.07 & 8.16  & 9.35  & 6.87  & 8.32  & \textbf{96.91} \\
		EZ-VSL(144k)~\cite{mo2022localizing}    & 10.92 & 12.22 & 6.97  & 8.34  & 5.80  & 7.42  & 96.47 \\
		EZ-VSL(full)~\cite{mo2022localizing}    & 12.17 & 13.38 & 7.67  & 8.91  & 6.96  & 8.40  & 95.43 \\
		SSL-TIE(144k)~\cite{liu2022exploiting}   & 13.10 & 14.23 & 5.19  & 6.76  & 5.50  & 7.12  & 90.82 \\
		TAVLO(10k)~\cite{choi2025s}     & 13.42 & 14.08 & 14.13 & 14.52 & 12.08 & 12.69 & 91.18 \\
		\midrule
		SlotAVS (Ours)        & \textbf{58.56} & \textbf{58.91} & \textbf{47.65} & \textbf{47.86} & \textbf{46.90} & \textbf{47.41} & 88.00 \\
		\bottomrule
	\end{tabular*}
	\vspace{2pt}
	\begin{flushleft}
	\textit{Note:} The baseline methods~\cite{mo2022closer,mo2022localizing, liu2022exploiting, choi2025s} on the AVATAR benchmark are trained using weaker supervision compared to SlotAVS. While this prevents a strictly fair benchmarking comparison, these results are included exclusively to demonstrate the strong zero-shot generalization and robust spatial grounding capabilities of our framework on unseen domains.
	\end{flushleft}
\end{table}

\subsection{Analysis of Architectural Design Choices}
\label{sec:ablation}
We ablate the core components of SlotAVS to isolate our algorithmic contributions from backbone representation capacity. To evaluate explicit disentanglement, we test these variants across both single source (S4) and multi-source (MS3) splits.

\begin{table}[h]
	\caption{\textbf{Ablation on object-centric Routing and Competitive Binding.} Evaluation of core architectural components on the S4 and MS3 splits. Higher values indicate better performance.}
	\label{tab:ablation-core}
	\centering
	\small
	\renewcommand{\arraystretch}{1.1}
	\setlength{\tabcolsep}{8pt}
	\begin{tabular}{lcccc}
		\toprule
		\multirow{2}{*}{Architecture Variant} & \multicolumn{2}{c}{S4 Split} & \multicolumn{2}{c}{MS3 Split} \\
		\cmidrule(lr){2-3} \cmidrule(lr){4-5}
		& $\mathcal{M}_J$ & $\mathcal{M}_F$ & $\mathcal{M}_J$ & $\mathcal{M}_F$ \\
		\midrule
		Global Audio (w/o Slots) & 86.9 & 0.928 & 69.2 & 0.769 \\ 
		Standard Cross Attention & 86.8 & \textbf{0.932} & 70.3 & \textbf{0.831} \\ 
		\midrule
		SlotAVS (Ours)           & \textbf{87.1} & 0.923 & \textbf{71.4} & 0.811 \\
		\bottomrule
	\end{tabular}
\end{table}

\noindent\textbf{Necessity of object-centric Routing:} To validate our core motivation, we bypass bi-modal slot attention and condition the spatial attention block directly on global, pooled audio features. As shown in Table~\ref{tab:ablation-core}, this variant performs well in the S4 split ($\mathcal{M}_J = 86.9$), confirming that global audio suffices for single dominant objects. However, performance degrades on the MS3 split, with $\mathcal{M}_J$ dropping to $69.2$ and boundary accuracy ($\mathcal{M}_F$) dropping sharply to $0.769$. Global audio inevitably entangles concurrent sources, causing merged masks. This confirms object-specific routing is structurally required for multi-source segmentation.

\noindent\textbf{Effect of Competitive Binding:} The attention mechanism's mathematical formulation is critical for true disentanglement. We ablate our competitive bottleneck by shifting softmax normalization from the slot dimension (\texttt{dim=1}) to the spatial dimension (\texttt{dim=2}) and removing population normalization, reverting to standard independent cross attention. While this variant yields a high contour accuracy ($\mathcal{M}_F = 0.831$) on MS3, this stems directly from query collapse: without a zero sum constraint, independent queries redundantly converge and over-fit to the most visually salient object's boundaries. This redundant focus sacrifices true multi-source separation, yielding a lower region similarity ($\mathcal{M}_J = 70.3$). Conversely, SlotAVS's competitive routing forces overlapping masks to distribute spatial probability mass. While this slightly softens intersecting boundaries ($\mathcal{M}_F = 0.811$), it promotes the instance level disentanglement necessary for robust localization ($\mathcal{M}_J = 71.4$).

\begin{table}[h]
	\caption{\textbf{Ablation on Asymmetric Foundation Adaptation:} Impact of applying LoRA to different modalities. The visual only adaptation (our main setting) provides the optimal balance of routing adaptability and semantic stability.}
	\label{tab:ablation-lora}
	\centering
	\small
	\renewcommand{\arraystretch}{1.1}
	\setlength{\tabcolsep}{8pt}
	\begin{tabular}{lcccc}
		\toprule
		\multirow{2}{*}{Adaptation Strategy} & \multicolumn{2}{c}{S4 Split} & \multicolumn{2}{c}{MS3 Split} \\
		\cmidrule(lr){2-3} \cmidrule(lr){4-5}
		& $\mathcal{M}_J$ & $\mathcal{M}_F$ & $\mathcal{M}_J$ & $\mathcal{M}_F$ \\
		\midrule
		Audio LoRA Only & 82.58 & 0.885 & 60.79 & 0.679 \\ 
		Both LoRA (Audio + Visual) & 86.95 & 0.925 & 65.72 & 0.768 \\ 
		None (Completely Frozen) & 85.11 & 0.912 & 67.07 & 0.790 \\ 
		\midrule
		Visual LoRA Only (Ours) & \textbf{87.10} & \textbf{0.923} & \textbf{71.40} & \textbf{0.811} \\
		\bottomrule
	\end{tabular}
\end{table}

\noindent\textbf{Asymmetric Foundation Adaptation:} 
Table~\ref{tab:ablation-lora} details the impact of our asymmetric LoRA strategy. When applying LoRA exclusively to the audio encoder, performance collapses severely on the MS3 split ($\mathcal{M}_J = 60.79$). This occurs because fine tuning the CLAP backbone destroys its pre trained alignment with the frozen text encoder. This semantic drift breaks the absolute reference anchor required by our Tri Modal Contrastive Feature Distillation ($\mathcal{L}_{CFD}$), leading to poor multi-source disentanglement. Tuning both modalities improves performance but still suffers from this acoustic drift. Conversely, completely freezing both backbones ($\mathcal{M}_J = 67.07$) preserves acoustic stability but prevents the visual backbone from learning the audio-conditioned spatial routing mechanisms demanded by the ACSA module. By applying LoRA strictly to the visual modality, we allow the spatial features to adapt to competitive object-centric routing while utilizing the frozen acoustic latent space as an immutable semantic anchor, achieving the optimal MS3 region similarity of $71.40$.

\noindent\textbf{Hyperparameter Sensitivity:} We evaluated the framework's robustness to the number of slots ($K$) and refinement iterations ($T_{iter}$). In the AVSBench dataset, a single sample contains up to three concurrent sounding objects. To successfully segregate these acoustic instances along with the visual background, a minimum of four slots ($K \ge 4$) is theoretically required. We empirically tested $K \in \{3, 4, 5, 6\}$. While all configurations with $K \ge 4$ yielded stable and comparable results, setting $K=5$ achieved the best overall performance and provides a necessary safety buffer for dense scenes. Furthermore, we ablated the number of iterations required for slot refinement. Scaling to $T_{iter} \in \{5, 7\}$ resulted in a very small performance gap compared to our baseline. Because these additional iterations increase computational cost with negligible segmentation gains, $T_{iter}=3$ was chosen as the final hyperparameter. Comprehensive data tables detailing these profiles, alongside ablations on the diversity loss weighting ($\mathcal{L}_{div}$), are provided in the supplementary material.

\begin{figure}[t]
	\centering
	\includegraphics[width=\linewidth]{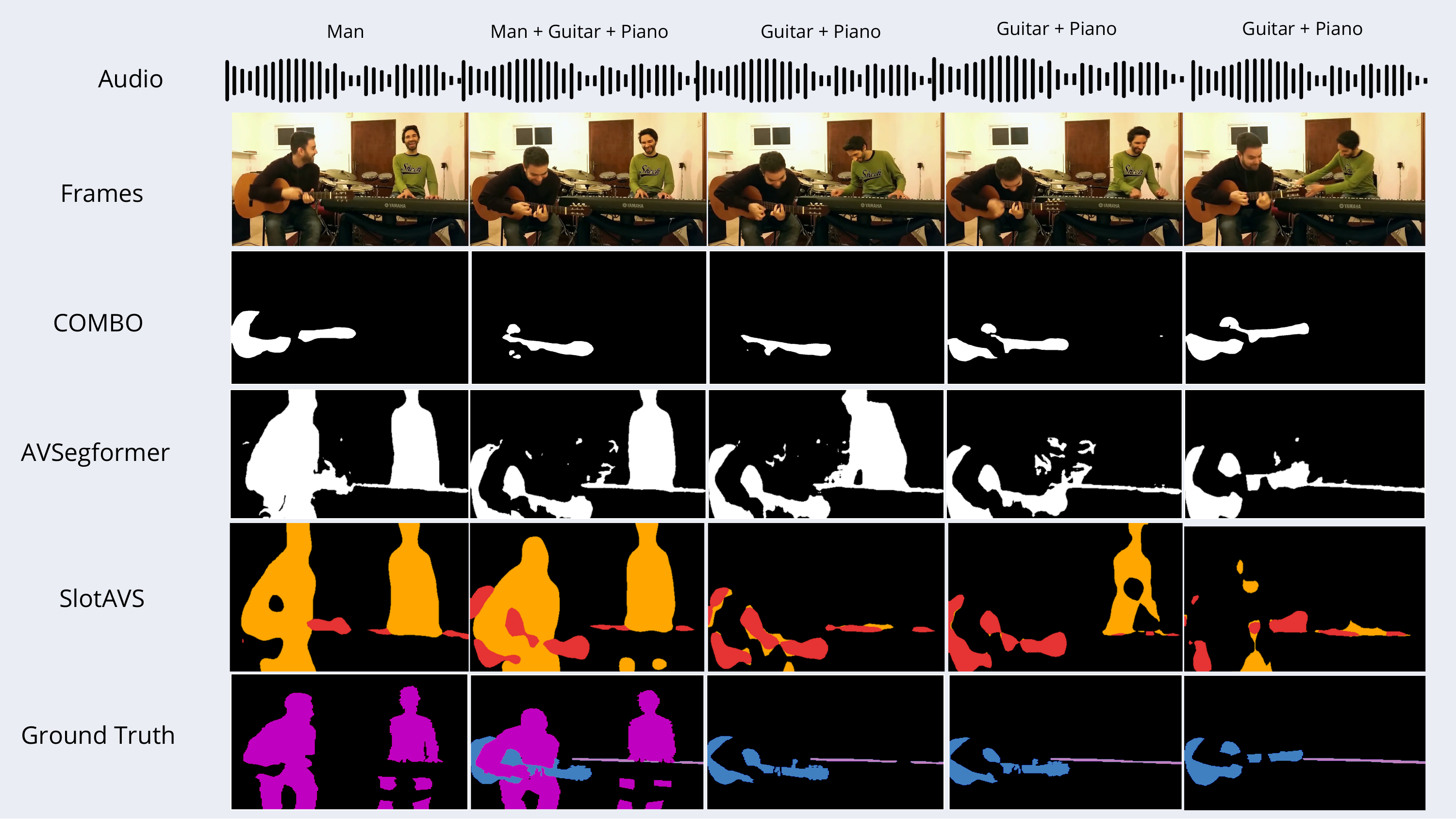}
	\caption{\textbf{Qualitative comparison in dynamic multi-source scenes.} SlotAVS distinctly segregates concurrent sounding objects (e.g., voice, guitar, piano) into separated instance masks, successfully mitigating the superposition and mask-merging effects commonly observed in standard global fusion approaches. \textbf{(Best viewed in color)}.}
	\label{fig:qualitative}
\end{figure}

\noindent\textbf{Qualitative Analysis:} We show segmentation in a dynamic multi-source scenario in Figure~\ref{fig:qualitative}. Under concurrent audio, e.g., simultaneous voice, guitar, and piano, recent baselines suffer from severe spatial superposition, merging distinct sounding objects into a single entangled mask. In contrast, SlotAVS leverages competitive binding to allocate independent, object-centric slots to each semantic entity. This explicit zero-sum spatial gating actively prevents mask merging, enforcing crisp instance boundaries, and robust disentanglement even among spatially proximate sources.

\subsection{Discussion}
\label{sec:discussion}
While SlotAVS mitigates cross-modal entanglement, it occasionally struggles with extreme intra class multi-source events, e.g., identical dogs barking. Nearly identical acoustic signatures hinder the temporal initialization and Contrastive Feature Distillation (CFD) loss from establishing distinct slot states. This ambiguity can propagate to the spatial Hungarian matcher, leading to slot instability.

Regarding architecture, we deliberately selected the DINOv3 foundation model over earlier backbones (e.g., PVT-v2) to leverage robust, generalizable spatial priors. While this limits direct backbone level comparisons, it is structurally essential for supporting our object-centric routing. 

Furthermore, input spatial resolution critically impacts performance. At a $224 \times 224$ resolution, the visual backbone produces only a $14 \times 14$ grid (196 tokens). With $K=5$ competing slots, overlapping objects in complex scenes share too few tokens, causing competitive slot assignment to degenerate. Scaling to $512 \times 512$ (1024 tokens) provides the necessary spatial granularity for clean disentanglement. However, operating on these high resolution tokens incurs a non trivial computational overhead, positioning SlotAVS as an accuracy prioritized framework. Future work will explore hierarchical or sparse attention paradigms to accelerate inference.

\section{Conclusion}
\label{sec:conclusion}
In this paper, we identified feature entanglement as a fundamental architectural bottleneck in multi-source Audio-Visual Segmentation. To address this, we proposed SlotAVS, shifting the paradigm from global cross-modal alignment to explicit, object-centric competitive binding. By initializing discrete slots from temporal audio sequences and enforcing strict zero-sum spatial competition, our architecture strongly discourages concurrent sounds from blending into unified masks. Coupled with rigorous instance-level spatial matching and tri-modal feature distillation, SlotAVS effectively isolates overlapping sound sources. Extensive experiments confirm that while foundation models provide necessary representational priors, explicit structural competition is critical for achieving robust, generalizable multi-source audio-visual scene understanding.

\section*{Acknowledgments}
The authors gratefully acknowledge Vehant Technologies Pvt. Ltd. for supporting this work.
\bibliography{egbib}
\end{document}